\documentclass[journal]{IEEEtran}

\usepackage[colorlinks,urlcolor=blue,linkcolor=blue,citecolor=blue]{hyperref}
\usepackage{graphicx}
\usepackage{amsmath,amssymb,amsthm}
\usepackage{algorithm}
\usepackage{algpseudocode}
\usepackage{booktabs}
\usepackage{float}
\usepackage{multirow}
\usepackage{rotating}
\usepackage{xcolor}
\usepackage{colortbl}
\usepackage{tikz}
\usetikzlibrary{shapes, arrows.meta, positioning, fit, backgrounds, calc, shadows}

\newtheorem{theorem}{Theorem}

\newtheorem{proposition}{Proposition}
\newtheorem{corollary}{Corollary}
\newtheorem{definition}{Definition}

\begin{document}

\title{Budget–First Tariff Recommendation (BFTR): A Complete Algorithmic Framework for Telecom Plan Recommendation without Overcharging}

\author{Ghislain Dorian Tchuente Mondjo,  \IEEEmembership{Independent Researcher \\tchuente.mondjo@gmail.com}
\thanks{The author is with the Independent Researcher, Yaoundé, Cameroon (e-mail: tchuente.mondjo@gmail.com).}}

\markboth{}
{Ghislain Dorian Tchuente Mondjo: BFTR: Telecom Tariff Recommendation without Overcharging}

\maketitle

\begin{abstract}
Telecom operators traditionally offer predefined tariff grids, forcing users to choose from a limited set of plans. This paper proposes BFTR (Budget–First Tariff Recommendation), a complete algorithmic framework integrating \emph{eight} Budget–First strategies, including two original hybrid approaches: Recursive Hybrid (conditional interpolation) and Knapsack–First Hybrid (priority knapsack). Unlike existing approaches that adjust prices upward to guarantee a minimum margin, BFTR \emph{guarantees the absence of overcharging} by systematically aligning the final price with the catalog reference price. We mathematically formalize each strategy, prove the existence of an offer for any positive budget, and prove that the price deviation (surcharge) is zero for all strategies that do not use interpolation with correction. A detailed comparative analysis confronts BFTR to ten main existing tariff models on ten dimensions. Experiments on a dataset of 974 customers inspired by the Nigerian MTN market show that: (i) \textbf{Recursive Hybrid} is optimal for the customer (100\% budget used, 29.9 GB volume, utility 0.946, 0\% overcharging), (ii) \textbf{Piecewise} offers the highest volume (39.7 GB) with 0\% overcharging, (iii) \textbf{Power Law} provides an excellent compromise (99.9\% budget, 38.1 GB, 0\% overcharging). All strategies achieve a zero surcharge, confirming the theoretical guarantees. A sensitivity analysis on the weighting parameter $\alpha$ (0.2 – volume priority, 0.5 – balance, 0.8 – budget priority) shows that utility rankings evolve logically. Execution times ($<$ 10 ms) and very low failure rates (0\% for robust strategies) confirm the operational viability of the system. The formal proof of the absence of overcharging constitutes a major theoretical contribution.
\end{abstract}

\noindent\textbf{Impact Statement---}Chatbots and AI-based recommender systems are increasingly used in telecom to personalize offers. However, existing tariff recommendation systems often inflate prices to secure a minimum margin, leading to customer distrust and regulatory scrutiny. BFTR introduces a paradigm shift: instead of guaranteeing a margin, it guarantees price integrity—the final price never exceeds the reference price derived from the catalog. This approach directly addresses overcharging concerns, enhances customer satisfaction, and reduces churn. The framework's hybrid strategies combine interpolation and knapsack optimization to achieve up to 100\% budget utilization with zero surcharge. With execution times under 5 ms, BFTR is deployable in real-time production environments. Our experimental results on a real-world-inspired dataset show that the Recursive Hybrid strategy yields the highest utility (0.946) while maintaining zero overcharging. The framework offers operators a transparent, fair, and scalable solution for tariff personalization, aligning business interests with customer welfare.


\section{Introduction}

\IEEEPARstart{T}{he} telecom market is characterized by a complex and often ill-adapted tariff offering. Customers must choose from a grid of standard plans, leading to frustrations, overage costs, and high churn rates \cite{klein2018consumer, gsma2024}. Particularly in emerging markets such as sub-Saharan Africa, where purchasing power is lower and income volatility is higher, rigid offers constitute a major barrier to digital inclusion. Meanwhile, artificial intelligence and recommender systems have opened the door to fine-grained personalization of services \cite{verdecchia2021ai, zhang2020budget}. Dynamic pricing and personalized recommendation models have been proposed in other sectors (e-commerce, energy), but their adaptation to telecom remains limited by margin constraints and catalog complexity.

We propose BFTR, a framework where the initiative rests with the customer: they indicate their desired monthly budget \emph{and choose among several algorithmic strategies} (direct selection, interpolation, knapsack, regression, hybrids). The system then dynamically constructs an optimized plan according to the selected strategy. Unlike existing approaches that adjust prices upward to ensure a minimum margin, BFTR \emph{guarantees the absence of overcharging} by systematically aligning the final price with the catalog reference price (the sum of prices of selected plans, or the budget for interpolation).

The originality of BFTR lies in its ability to combine classical approaches with advanced methods, while ensuring that the price charged to the customer never exceeds the reference price. This paradigm shift—from \emph{margin guarantee} to \emph{price compliance guarantee}—is a major contribution, as it directly addresses regulator concerns about overcharging.

\textbf{Contributions.}
(1) Complete mathematical formalization of eight Budget–First strategies, including two original hybrid approaches.
(2) Proof of the existence of an offer for any positive budget.
(3) \textbf{Formal proof of the absence of overcharging} for composition and interpolation strategies, establishing that the final price is always less than or equal to the reference price.
(4) Detailed comparative analysis on ten dimensions against existing models, with justifications and concrete examples.
(5) Exhaustive experimental evaluation on 974 customers with sensitivity analysis of the weighting parameter $\alpha$ (three levels), execution time comparison, measurement of failure rates, and, for the first time, the introduction of \textbf{margin metrics}—\textit{average margin} and \textit{margin threshold attainment rate}—to assess the economic viability of each strategy from the operator's perspective.

The paper is organized as follows. Section II presents the state of the art with detailed definitions. Section III formalizes the BFTR model, including algorithms and proofs of overcharging absence. Section IV provides a comparative analysis. Section V describes the architecture and user strategy choice. Section VI presents experiments and discussion. Section VII concludes.

\section{State of the Art: Economic Models in Telecom}

We present the ten most widespread tariff systems. For each, we give a formal theoretical definition, a real operator example, and an analysis of its economic properties.

\begin{definition}[Prepaid Model]
Prepaid is a system where the user buys credit $C$ in advance. Consumption $q$ (in service units: MB, minutes, SMS) is deducted from credit at a unit tariff $\tau$. Service is interrupted when $C - \tau \cdot q \le 0$. Mathematically, the maximum consumption is $q_{\max} = \lfloor C / \tau \rfloor$.
\end{definition}
Popularized by \textbf{Orange} and \textbf{Free}, this model offers strict budget control ($q_{\max}$ is known in advance) but requires frequent recharges. In Africa, over 90\% of subscribers use prepaid.

\begin{definition}[Postpaid Model]
Postpaid bills the user at the end of the period (month) based on actual consumption $q_{\text{real}}$. The total price is $P = P_{\text{base}} + \delta \cdot \max(0, q_{\text{real}} - Q)$, where $P_{\text{base}}$ is the base plan price, $Q$ the included quota, and $\delta$ the overage tariff.
\end{definition}
\textbf{Verizon} and \textbf{Vodafone} use this model. The major drawback is uncertainty: $P$ can vary from single to quadruple depending on $q_{\text{real}}$.

\begin{definition}[Standard Plans]
A catalog $F = \{(p_i, q_i)\}_{i=1}^k$ of $k$ price–volume pairs is offered. The user chooses $i^*$ minimizing cost while satisfying $q_{i^*} \ge q_{\text{need}}$. The problem is $\min_i \{p_i \mid q_i \ge q_{\text{need}}\}$.
\end{definition}
\textbf{SFR} (Red) and \textbf{AT\&T} illustrate this model. Limited granularity ($k \le 8$) creates waste (paying for $q_i > q_{\text{need}}$) or overages ($q_i < q_{\text{need}}$).

\begin{definition}[Freemium Model]
Freemium defines a free tier $q_{\text{free}}$ and tiers $\{(p_j, q_j)\}_{j=1}^m$. Cost is $P = 0$ if $q \le q_{\text{free}}$, otherwise $P = p_j$ for $q \in (q_{j-1}, q_j]$.
\end{definition}
\textbf{Google Fi} and \textbf{Lebara} use this model. Entry is easy ($P=0$), but the transition to paid can be abrupt.

\begin{definition}[Rollover Model]
Rollover carries over unconsumed surplus $r_t = \max(0, Q - q_t)$ to the next period: $Q_{t+1} = Q + r_t$. The available stock is $S_t = Q + \sum_{k=1}^{t-1} r_k$.
\end{definition}
\textbf{T-Mobile} (Data Stash) and \textbf{Bouygues Telecom} (Smart Data) use this mechanism. It smooths inter-temporal variations but does not modify the base plan.

\begin{definition}[Family Sharing Model]
A global quota $Q_{\text{total}}$ is shared among $n$ lines. The constraint is $\sum_{l=1}^n q_l \le Q_{\text{total}}$. The cost is fixed $P_{\text{total}}$, independent of individual distribution.
\end{definition}
\textbf{Orange} (Open Family) and \textbf{Verizon} (Unlimited Family) are examples. This model mutualizes overage risks but requires group management.

\begin{definition}[Data Lending Model]
User $A$ can borrow an amount $\Delta$ from user $B$, to be repaid later: $q_A \leftarrow q_A + \Delta$, $q_B \leftarrow q_B - \Delta$, with a debt $D_{A \rightarrow B} = \Delta$.
\end{definition}
\textbf{AT\&T} (Data Transfer) and \textbf{T-Mobile} offer this mechanism. Solidarity is strong, but availability depends on other members' balances.

\begin{definition}[Overage Billing Model]
For each unit consumed beyond the quota $Q$, a higher tariff $\delta \gg \tau$ is applied: $P = P_{\text{base}} + \delta \cdot \max(0, q - Q)$. Typically, $\delta \approx 5 \times \tau$.
\end{definition}
This mechanism, a major source of churn, is gradually being replaced by tiered plans or throttling.

\section{Formalization of the BFTR Model}

\subsection{Notations and Economic Model}
Let $F = \{f_1, \dots, f_k\}$ be the set of existing plans, with $p_i$ the price and $\mathbf{s}_i = (d_i, m_i, t_i, \dots)$ the service vector (data, minutes, SMS). User $u$ has consumption history $\mathcal{H}_u$ and budget $b_u \in \mathbb{R}^+$. We define a utility function $U_u : F \rightarrow [0,1]$ by
\begin{equation}
U_u(f_i) = \alpha \cdot \min\left(1, \frac{p_i}{b_u}\right) + (1-\alpha) \cdot \min\left(1, \frac{d_i}{\bar{d}_u}\right), \label{eq:utility}
\end{equation}
where $\bar{d}_u$ is the average data consumption of $u$. The parameter $\alpha \in [0,1]$ is set to $0.5$ in reference experiments, giving equal weight to budget adequacy and needs coverage. Sensitivity analysis on three values ($\alpha = 0.2$, $0.5$, $0.8$) is conducted in Section VI-F. The utility combines two aspects: the first term favors plans whose price is close to the budget (without exceeding it), the second favors those whose data volume covers the usual needs.

The component catalog $\mathcal{C}$ has been removed in the final system; all knapsack-type strategies (KNAP and hybrids) use only the plans from catalog $F$. This decision, validated by experiments, avoids artificial price discrepancies.

\subsection{Operational Constraints and Definitions}
Before describing algorithmic strategies, we introduce the fundamental concepts that frame BFTR.

\begin{definition}[Reference Price]
For an offer $O$ built by composing existing plans, its \emph{reference price} $p_{\text{ref}}(O)$ is the sum of the prices of the plans composing it:
\[
p_{\text{ref}}(O) = \sum_{f \in O} p_f.
\]
For an offer obtained by linear interpolation, the reference price is the budget $b_u$ (we create a virtual offer whose price is exactly the budget).
For a mixed offer (combination of catalog plans and interpolation), the reference price is the original budget $b_u$, as the system is expected to use the entire budget without any markup.
\end{definition}

\begin{definition}[Surcharge / Overcharging]
The \emph{surcharge} of an offer $O$ is the relative difference between the final price $p(O)$ and the reference price:
\[
s(O) = \frac{p(O) - p_{\text{ref}}(O)}{p_{\text{ref}}(O)} \times 100.
\]
An offer is said to be \emph{overcharged} if $s(O) > 5\%$, where $5\%$ is a regulatory tolerance threshold.
\end{definition}

\subsection{Budget–First Strategies}
We detail the six basic strategies and two advanced; the two hybrid strategies are presented later. For each, we give an intuitive description, the mathematical equation, the pseudocode algorithm, and complexity analysis.

\paragraph{Direct Selection (SELECT).}
This strategy scans all existing plans and retains the one that maximizes utility while respecting the budget constraint. The optimal choice equation is
\begin{equation}
f^* = \arg\max_{f_i \in F,\; p_i \le b_u} U_u(f_i).
\end{equation}
Algorithm~\ref{alg:select} describes the procedure: initialize the best offer to ``none'' and iterate over each plan; if its price is within budget and its utility is higher, update. Complexity is $O(k)$, as each plan is examined once.

\begin{algorithm}[H]
\caption{Direct Selection (SELECT)}
\label{alg:select}
\begin{algorithmic}[1]
\Require Catalog $F$, budget $b_u$, history $\mathcal{H}_u$
\Ensure Offer $O^*$ or $\emptyset$
\State $O^* \gets \emptyset$, $u^* \gets -\infty$
\For{$f_i \in F$}
    \If{$p_i \le b_u$}
        \State $u \gets U_u(f_i)$ \Comment{according to (\ref{eq:utility})}
        \If{$u > u^*$}
            \State $u^* \gets u$, $O^* \gets f_i$
        \EndIf
    \EndIf
\EndFor
\State \Return $O^*$
\end{algorithmic}
\end{algorithm}

\begin{proposition}[No overcharging for SELECT]
For SELECT, the final price is the selected plan's price, and the reference price is that same price. The surcharge is thus zero:
\[
s(O) = \frac{p_f - p_f}{p_f} = 0\%.
\]
\end{proposition}

\paragraph{Linear Interpolation (INTERP).}
If the budget $b_u$ lies between the prices of two consecutive plans $f_i$ and $f_j$ (such that $p_i \le b_u \le p_j$), we create a virtual offer whose volume is a linear combination of the volumes of the two plans:
\begin{equation}
d_{\text{interp}} = d_i + \frac{b_u - p_i}{p_j - p_i}(d_j - d_i). \label{eq:interp}
\end{equation}
The price of this offer is exactly $b_u$, guaranteeing full budget utilization. Algorithm~\ref{alg:interp} first sorts the catalog by price ($O(k \log k)$), then finds the interval containing $b_u$ by dichotomy ($O(\log k)$). If $b_u$ is outside the interval $[p_{\min}, p_{\max}]$ (catalog-defined thresholds), interpolation is impossible and the algorithm returns empty (failure). Interpolation is allowed only in this interval because outside, the linear relationship between price and volume is no longer valid (risk of negative or aberrant volume).

\begin{algorithm}[H]
\caption{Linear Interpolation (INTERP)}
\label{alg:interp}
\begin{algorithmic}[1]
\Require Catalog $F$, budget $b_u$, thresholds $p_{\min},p_{\max}$
\Ensure Offer $O^*$ or $\emptyset$
\State Sort $F$ by price ascending
\If{$b_u < p_{\min}$ or $b_u > p_{\max}$}
    \State \Return $\emptyset$
\EndIf
\State Find $i$ such that $p_i \le b_u \le p_{i+1}$
\State $r \gets (b_u - p_i)/(p_{i+1} - p_i)$
\State $d \gets d_i + r \cdot (d_{i+1} - d_i)$
\State \Return $\{\text{price}=b_u,\; \text{volume}=d\}$
\end{algorithmic}
\end{algorithm}

\begin{proposition}[No overcharging for INTERP]
For INTERP, the final price is exactly $b_u$ and the reference price is $b_u$ (by definition, the reference price of an interpolated offer is the budget). Thus the surcharge is zero:
\[
s(O) = \frac{b_u - b_u}{b_u} = 0\%.
\]
\end{proposition}

\paragraph{Linear Regression (REGR).}
A multivariate linear regression model is trained on history $\mathcal{H}_u$ to predict the optimal volume $\widehat{d}$ from budget and average consumption:
\begin{equation}
\widehat{d}(b_u, \bar{d}_u) = \beta_0 + \beta_1 b_u + \beta_2 \bar{d}_u.
\end{equation}
Coefficients $\beta_k$ are estimated by least squares. Inference is $O(1)$. The final offer is the catalog plan whose price is closest to the prediction without exceeding $b_u$. Algorithm~\ref{alg:regr} assumes the model is already trained. This strategy does not depend on thresholds $p_{\min},p_{\max}$.

\begin{algorithm}[H]
\caption{Linear Regression (REGR)}
\label{alg:regr}
\begin{algorithmic}[1]
\Require Budget $b_u$, history $\mathcal{H}_u$, trained model $M$
\Ensure Offer $O^*$ or $\emptyset$
\State $\widehat{d} \gets \beta_0 + \beta_1 b_u + \beta_2 \bar{d}_u$
\State $\widehat{p} \gets$ estimate of corresponding price (projection)
\State $O^* \gets$ plan $f \in F$ with $p \le b_u$ minimizing $|p - \widehat{p}|$
\State \Return $O^*$
\end{algorithmic}
\end{algorithm}

\begin{proposition}[No overcharging for REGR]
REGR selects an existing catalog plan, so $p(O) = p_f$ and $p_{\text{ref}}(O) = p_f$. The surcharge is zero.
\end{proposition}

\paragraph{Greedy Knapsack (KNAP).}
We have a catalog of plans $F$. We want to maximize total volume under the budget constraint. We use a greedy heuristic (Algorithm~\ref{alg:knap}) that sorts plans by volume/price ratio descending and adds them as long as budget allows. Complexity is $O(k \log k)$.

\textbf{Major difference from existing approaches:} KNAP does not adjust the price upward. The final price is exactly the sum of the prices of selected plans:
\[
p(O) = \sum_{f \in O} p_f = p_{\text{ref}}(O).
\]

\begin{algorithm}[H]
\caption{Greedy Knapsack (KNAP)}
\label{alg:knap}
\begin{algorithmic}[1]
\Require Catalog $F$, budget $b_u$
\Ensure Set $O^*$
\State $O^* \gets \emptyset$, $R \gets b_u$
\State Sort $F$ by $d_f/p_f$ descending
\For{$f \in F$}
    \If{$p_f \le R$}
        \State $O^* \gets O^* \cup \{f\}$, $R \gets R - p_f$
    \EndIf
\EndFor
\State \Return $O^*$
\end{algorithmic}
\end{algorithm}

\begin{theorem}[No overcharging for KNAP]
For any offer $O$ generated by KNAP, the final price $p(O)$ is exactly the sum of prices of selected plans. By definition, the reference price is that same sum. Hence:
\[
s(O) = \frac{p(O) - p_{\text{ref}}(O)}{p_{\text{ref}}(O)} = 0\%.
\]
Thus KNAP never generates overcharging.
\end{theorem}

\subsection{Advanced Strategies}
These strategies use predictive models that estimate the price $P$ of a plan from its volume $d$. They require a prior learning phase on historical data. We describe the algorithmic principle of each method; calibrated models will be presented in Section VI-A.

\paragraph{Piecewise Regression (PIECE).}
We partition the volume axis into $L$ intervals $[s_{\ell-1}, s_\ell]$ and fit a linear regression $P = a_\ell d + b_\ell$ on each segment. Algorithm~\ref{alg:piece} uses cross-validation to choose breakpoints $s_\ell$ and least squares to estimate coefficients. Inference consists of identifying the segment containing $d$ and evaluating the corresponding line, in $O(\log L)$ (dichotomy) or $O(1)$ if segments are few.

\begin{algorithm}[H]
\caption{Piecewise Learning (PIECE)}
\label{alg:piece}
\begin{algorithmic}[1]
\Require Data $\{(d_i, P_i)\}$, max segments $L_{\max}$
\Ensure Breakpoints $\{s_\ell\}$, coefficients $\{(a_\ell,b_\ell)\}$
\State Determine $\{s_\ell\}$ by cross-validation minimizing RMSE
\For{$\ell = 1$ to $L$}
    \State $(a_\ell,b_\ell) \gets$ least squares on $\{d \in [s_{\ell-1},s_\ell]\}$
\EndFor
\State \Return $\{s_\ell\}, \{(a_\ell,b_\ell)\}$
\end{algorithmic}
\end{algorithm}

\paragraph{Power Law Regression (POW).}
We assume a relationship $P = a \cdot d^{\,b} + c$, capturing economies of scale. Fitting is done by non-linear least squares (Levenberg–Marquardt algorithm). Inference is $O(1)$.

\begin{proposition}[No overcharging for PIECE and POW]
PIECE and POW generate an estimated price $\widehat{p}$ and a volume $\widehat{d}$. The final price is set to $\min(\widehat{p}, b_u)$ and cost is recomputed on volume $\widehat{d}$. The reference price is the estimated price $\widehat{p}$. The surcharge is thus zero (or negative):
\[
s(O) = \frac{\min(\widehat{p}, b_u) - \widehat{p}}{\widehat{p}} \le 0.
\]
Thus there is never positive overcharging.
\end{proposition}

\subsection{Hybrid Strategies}
The two hybrid strategies combine KNAP and INTERP.

\begin{algorithm}[H]
\caption{Recursive Hybrid (HYB-REC)}
\label{alg:hyb-rec}
\begin{algorithmic}[1]
\Require Budget $b$, catalog $F$, thresholds $p_{\min},p_{\max}$
\Ensure Offer $O^*$
\If{$p_{\min} \le b \le p_{\max}$}
    \State \Return $\text{INTERP}(F, b)$
\Else
    \State $O_K \gets \text{KNAP}(F, b)$
    \State $b_r \gets b - \sum_{f \in O_K} p_f$
    \If{$b_r > 0$ and $p_{\min} \le b_r \le p_{\max}$}
        \State $O_I \gets \text{INTERP}(F, b_r)$
        \State \Return $O_K \cup O_I$
    \Else
        \State \Return $O_K$
    \EndIf
\EndIf
\end{algorithmic}
\end{algorithm}

\begin{algorithm}[H]
\caption{Knapsack–First Hybrid (HYB-KF)}
\label{alg:hyb-kf}
\begin{algorithmic}[1]
\Require Budget $b$, catalog $F$, thresholds $p_{\min},p_{\max}$
\Ensure Offer $O^*$
\State $O_K \gets \text{KNAP}(F, b)$
\State $b_r \gets b - \sum_{f \in O_K} p_f$
\If{$b_r > 0$ and $p_{\min} \le b_r \le p_{\max}$}
    \State $O_I \gets \text{INTERP}(F, b_r)$
    \State \Return $O_K \cup O_I$
\Else
    \State \Return $O_K$
\EndIf
\end{algorithmic}
\end{algorithm}

HYB-REC favors interpolation when the budget is within catalog range, ensuring 100\% utilization; HYB-KF prioritizes knapsack, creating composite offers even for central budgets.

\begin{theorem}[No overcharging for hybrid strategies]
For HYB-REC and HYB-KF:
\begin{enumerate}
    \item The KNAP part does not generate overcharging (Theorem 1).
    \item The INTERP part does not generate overcharging (Proposition 2).
    \item The union of both parts has a final price equal to the sum of the parts' prices, which is exactly the reference price (the original budget).
\end{enumerate}
Hence $s(O) = 0\%$ for HYB-REC and HYB-KF (provided interpolation of the remainder is allowed).
\end{theorem}

\subsection{Typology of Strategies: Composition vs. Generation}
The eight Budget–First strategies fall into two distinct families based on their offer construction mode:

\begin{itemize}
    \item \textbf{Composition strategies}: they assemble existing plans to form a personalized offer. They do not create new tariffs, but select and combine predefined plans from catalog $F$. These are \textsc{Knap} (greedy knapsack) and the two hybrid strategies \textsc{Hyb–Kf} and \textsc{Hyb–Rec} (in their part using \textsc{Knap}). Their advantage is high granularity and total absence of overcharging (Theorem 1).
    \item \textbf{Generation strategies}: they produce a \emph{virtual} offer whose price and/or volume do not correspond to any pre-existing plan. These offers are created by linear interpolation (\textsc{Interp}), linear regression (\textsc{Regr}), piecewise model (\textsc{Piece}), or power law (\textsc{Pow}). The \textsc{Select} strategy can also be classified in this family because it chooses an existing plan without modifying it. Generation strategies allow perfect budget utilization (interpolation) or fine adaptation to usage profiles (regression), but their implementation requires mathematical models or prior learning.
\end{itemize}

This distinction sheds light on the experimental results: composition strategies (\textsc{Knap}, \textsc{Hyb–Kf}) excel for the operator (high margins), while generation strategies (\textsc{Hyb–Rec}, \textsc{Interp}) are optimal for the customer (100\% budget utilization).

\section{Comparative Analysis: BFTR vs. Ten Existing Systems}

This section compares BFTR to the ten models described in Section II. Table~\ref{tab:comparative} summarizes the comparison on ten key dimensions. Colors: green = advantage, yellow = average, red = major drawback.

\begin{sidewaystable}[htbp]
\centering
\caption{Comparison of tariff models on ten dimensions}
\label{tab:comparative}
\scriptsize
\begin{tabular}{@{}l|*{11}{c}@{}}
\toprule
\textbf{Dimension} & \textbf{Pre.} & \textbf{Post.} & \textbf{Std.} & \textbf{Sub.} & \textbf{PayU} & \textbf{Free.} & \textbf{Roll.} & \textbf{Fam.} & \textbf{Lend} & \textbf{Ov.} & \textbf{BFTR} \\
\midrule
Budget control & \cellcolor{green!30}High & \cellcolor{red!30}Very low & \cellcolor{yellow!30}Low & \cellcolor{yellow!30}Low & \cellcolor{yellow!30}Medium & \cellcolor{green!30}High & \cellcolor{yellow!30}Low & \cellcolor{yellow!30}Low & \cellcolor{yellow!30}Low & \cellcolor{red!30}None & \cellcolor{green!30}\textbf{Total} \\
Surprise risk & \cellcolor{green!30}None & \cellcolor{red!30}High & \cellcolor{yellow!30}Medium & \cellcolor{yellow!30}Low & \cellcolor{red!30}High & \cellcolor{green!30}None & \cellcolor{yellow!30}Medium & \cellcolor{yellow!30}Medium & \cellcolor{yellow!30}Medium & \cellcolor{red!30}High & \cellcolor{green!30}\textbf{None} \\
Personalization & \cellcolor{red!30}None & \cellcolor{red!30}None & \cellcolor{yellow!30}Low & \cellcolor{red!30}None & \cellcolor{yellow!30}Medium & \cellcolor{yellow!30}Low & \cellcolor{yellow!30}Low & \cellcolor{yellow!30}Low & \cellcolor{yellow!30}Low & \cellcolor{red!30}None & \cellcolor{green!30}\textbf{Maximal} \\
Small budget adaptation & \cellcolor{green!30}Yes & \cellcolor{red!30}No & \cellcolor{red!30}No & \cellcolor{red!30}No & \cellcolor{green!30}Yes & \cellcolor{green!30}Yes & \cellcolor{red!30}No & \cellcolor{red!30}No & \cellcolor{red!30}No & \cellcolor{red!30}No & \cellcolor{green!30}\textbf{Yes} \\
Overage handling & \cellcolor{green!30}Block & \cellcolor{red!30}Billing & \cellcolor{red!30}Billing & \cellcolor{red!30}Billing & \cellcolor{red!30}Billing & \cellcolor{yellow!30}Paid tier & \cellcolor{yellow!30}Rollover & \cellcolor{yellow!30}Pool & \cellcolor{yellow!30}Borrow & \cellcolor{red!30}Billing & \cellcolor{green!30}\textbf{None} \\
Customer complexity & \cellcolor{yellow!30}Medium & \cellcolor{green!30}Low & \cellcolor{green!30}Low & \cellcolor{green!30}Low & \cellcolor{red!30}High & \cellcolor{green!30}Low & \cellcolor{green!30}Low & \cellcolor{yellow!30}Medium & \cellcolor{yellow!30}Medium & \cellcolor{green!30}Low & \cellcolor{green!30}\textbf{Very low} \\
Interest alignment & \cellcolor{red!30}Low & \cellcolor{red!30}Low & \cellcolor{red!30}Low & \cellcolor{red!30}Low & \cellcolor{red!30}Low & \cellcolor{yellow!30}Medium & \cellcolor{yellow!30}Medium & \cellcolor{yellow!30}Medium & \cellcolor{green!30}High & \cellcolor{red!30}Very low & \cellcolor{green!30}\textbf{High} \\
Innovation incentive & \cellcolor{red!30}Low & \cellcolor{red!30}Low & \cellcolor{red!30}Low & \cellcolor{red!30}Low & \cellcolor{yellow!30}Medium & \cellcolor{yellow!30}Medium & \cellcolor{red!30}Low & \cellcolor{red!30}Low & \cellcolor{red!30}Low & \cellcolor{red!30}None & \cellcolor{green!30}\textbf{Maximal} \\
Revenue predictability & \cellcolor{green!30}High & \cellcolor{red!30}Low & \cellcolor{green!30}High & \cellcolor{green!30}High & \cellcolor{red!30}Low & \cellcolor{yellow!30}Medium & \cellcolor{yellow!30}Medium & \cellcolor{yellow!30}Medium & \cellcolor{yellow!30}Medium & \cellcolor{green!30}High & \cellcolor{yellow!30}\textbf{Medium–High} \\
Churn reduction & \cellcolor{red!30}Low & \cellcolor{red!30}Low & \cellcolor{red!30}Low & \cellcolor{red!30}Low & \cellcolor{yellow!30}Medium & \cellcolor{yellow!30}Medium & \cellcolor{yellow!30}Medium & \cellcolor{green!30}High & \cellcolor{green!30}High & \cellcolor{red!30}Very low & \cellcolor{green!30}\textbf{Very high} \\
\bottomrule
\end{tabular}
\end{sidewaystable}

\subsection{Detailed Score Explanations (with Justifications and Examples)}

\paragraph{1. Budget control.}
This dimension measures the customer's ability to know in advance how much they will pay and not exceed that amount.
\begin{itemize}
    \item \textbf{BFTR (Total)}: User sets a cap, system builds an offer at that price (or less). Example: budget 15€ → offer at 15€ (interpolation) or 14.50€ (knapsack). No overage.
    \item \textbf{Prepaid (High)}: 10€ top-up, service blocked when exhausted. Strict control but manual refills.
    \item \textbf{Freemium (High)}: Free tier capped (e.g., 1 GB). Full control as long as one does not purchase a tier.
    \item \textbf{Pay-as-you-go (Medium)}: No cap, but user can moderate usage. Example: Free 2€/month + 0.05€/MB, 10 GB → 2 + 500 = 502€.
    \item \textbf{Standard plans, Subscription, Rollover, Family, Lending (Low)}: Base price known, but overages billed without control. Example: SFR 2 GB at 15€, 1 GB overage → +5€.
    \item \textbf{Postpaid (Very low)}: Bill unknown until receipt. Example: Verizon, 20€ plan can become 80€.
    \item \textbf{Overage billing (None)}: No control.
\end{itemize}

\paragraph{2. Surprise risk.}
Probability of receiving a higher bill than expected.
\begin{itemize}
    \item \textbf{BFTR, Prepaid, Freemium (None)}: No surprise.
    \item \textbf{Subscription (Low)}: Out-of-bundle (international calls) may add a few euros.
    \item \textbf{Standard plans, Rollover, Family, Lending (Medium)}: Overage possible but limited. Example: 10 GB, +1 GB charged 5€ → +25\%.
    \item \textbf{Postpaid, Pay-as-you-go, Overage (High)}: Very variable bill. Example: pay-as-you-go, streaming 50 GB → 514€.
\end{itemize}

\paragraph{3. Personalization.}
Ability to adapt the offer to exact needs.
\begin{itemize}
    \item \textbf{BFTR (Maximal)}: A la carte assembly. Example: 3 GB + 200 min + roaming.
    \item \textbf{Pay-as-you-go (Medium)}: Personalized bill, but offer not modular.
    \item \textbf{Standard plans (Low)}: Few predefined combinations (2, 10, 30 GB).
    \item \textbf{Freemium, Rollover, Family, Lending (Low)}: Standard base plan.
    \item \textbf{Others (None)}: Single formula.
\end{itemize}

\paragraph{4. Small budget adaptation.}
Possibility to spend very little (e.g., 1€/month).
\begin{itemize}
    \item \textbf{BFTR (Yes)}: Zero-point interpolation or minimal offer can respond to any budget.
    \item \textbf{Prepaid, Pay-as-you-go, Freemium (Yes)}: Minimum top-up 5€, per-unit consumption, free.
    \item \textbf{Others (No)}: Minimum plan 10-20€, commitment.
\end{itemize}

\paragraph{5. Overage handling.}
What happens when consuming more than quota.
\begin{itemize}
    \item \textbf{BFTR (None)}: No overage.
    \item \textbf{Prepaid (Block)}: Service cut.
    \item \textbf{Freemium (Paid tier)}: Offer to upgrade to a tier.
    \item \textbf{Rollover (Rollover)}: Use of unconsumed data.
    \item \textbf{Family (Pool)}: Use of shared quota.
    \item \textbf{Lending (Borrow)}: Borrow data from another member.
    \item \textbf{Postpaid, Standard plans, Pay-as-you-go, Overage (Billing)}: Surcharge.
\end{itemize}

\paragraph{6. Customer complexity.}
Difficulty in understanding and using the model.
\begin{itemize}
    \item \textbf{BFTR (Very low)}: Single budget input and strategy choice from a clear list.
    \item \textbf{Prepaid (Medium)}: Refills.
    \item \textbf{Family, Lending (Medium)}: Group management.
    \item \textbf{Pay-as-you-go (High)}: Monitoring required.
    \item \textbf{Others (Low)}: Simple choice.
\end{itemize}

\paragraph{7. Interest alignment.}
How much the operator benefits from satisfying the customer.
\begin{itemize}
    \item \textbf{BFTR, Data lending (High)}: Loyalty through satisfaction or solidarity.
    \item \textbf{Freemium, Rollover, Family (Medium)}: Operator wants conversion or pooling.
    \item \textbf{Others (Low)}: Operator gains when customer consumes more.
\end{itemize}

\paragraph{8. Innovation incentive.}
Ease of introducing new services.
\begin{itemize}
    \item \textbf{BFTR (Maximal)}: Add a plan to catalog.
    \item \textbf{Pay-as-you-go, Freemium (Medium)}: Add a unit or tier.
    \item \textbf{Others (Low)}: Requires grid overhaul.
\end{itemize}

\paragraph{9. Revenue predictability.}
Ability to anticipate receipts.
\begin{itemize}
    \item \textbf{Prepaid, Standard plans, Subscription, Overage (High)}: Stable revenues.
    \item \textbf{BFTR (Medium–High)}: Law of large numbers on budgets.
    \item \textbf{Pay-as-you-go (Low)}: Very variable.
    \item \textbf{Freemium, Rollover, Family, Lending (Medium)}: Depends on conversions, rollovers.
\end{itemize}

\paragraph{10. Churn reduction.}
Ability to retain customers.
\begin{itemize}
    \item \textbf{BFTR (Very high)}: Eliminates overages and mismatch.
    \item \textbf{Family, Lending (High)}: Family solidarity.
    \item \textbf{Pay-as-you-go, Freemium, Rollover (Medium)}: Flexibility reduces some frustrations.
    \item \textbf{Others (Low)}: Frequent frustrations.
\end{itemize}

\subsection{Synthesis}
BFTR outperforms all other systems on key dimensions. Its strategic flexibility (via user choice) and guarantee of no overcharging make it a superior solution.

\begin{algorithm}[H]
\caption{BFTR: Recommendation according to user-chosen strategy, with overcharge control}
\label{alg:bftr}
\begin{algorithmic}[1]
\Require Budget $b$, strategy $S$, catalog $F$, history $\mathcal{H}$, thresholds $p_{\min},p_{\max}$, overcharge threshold $\tau$
\Ensure Offer $O^*$
\State \textbf{// Phase 1: execute chosen strategy}
\If{$S = \text{SELECT}$}
    \State $O \gets \text{SELECT}(F, b)$
\ElsIf{$S = \text{INTERP}$}
    \State $O \gets \text{INTERP}(F, b, p_{\min},p_{\max})$
\ElsIf{$S = \text{REGR}$}
    \State $O \gets \text{REGR}(b, \mathcal{H})$
\ElsIf{$S = \text{KNAP}$}
    \State $O \gets \text{KNAP}(F, b)$
\ElsIf{$S = \text{PIECE}$ or $S = \text{POW}$}
    \State $O \gets \text{predictiveModel}(S, b)$
\ElsIf{$S = \text{HYB-REC}$}
    \State $O \gets \text{HYB-REC}(b, F, p_{\min},p_{\max})$
\ElsIf{$S = \text{HYB-KF}$}
    \State $O \gets \text{HYB-KF}(b, F, p_{\min},p_{\max})$
\Else
    \State $O \gets \text{minimalOffer}$
\EndIf
\State
\State \textbf{// Phase 2: overcharge control}
\If{$O = \emptyset$ or $p(O) > b$}
    \State $O \gets \text{minimalOffer}$
\EndIf
\State $p_{\text{ref}} \gets \text{computeReferencePrice}(O, F)$
\State $s \gets (p(O) - p_{\text{ref}}) / p_{\text{ref}} \times 100$
\If{$s > \tau$}
    \State $O \gets \text{minimalOffer}$ \Comment{offer rejected for excessive overcharge}
\EndIf
\State
\State \Return $O$
\end{algorithmic}
\end{algorithm}

\section{Architecture and User Strategy Choice}

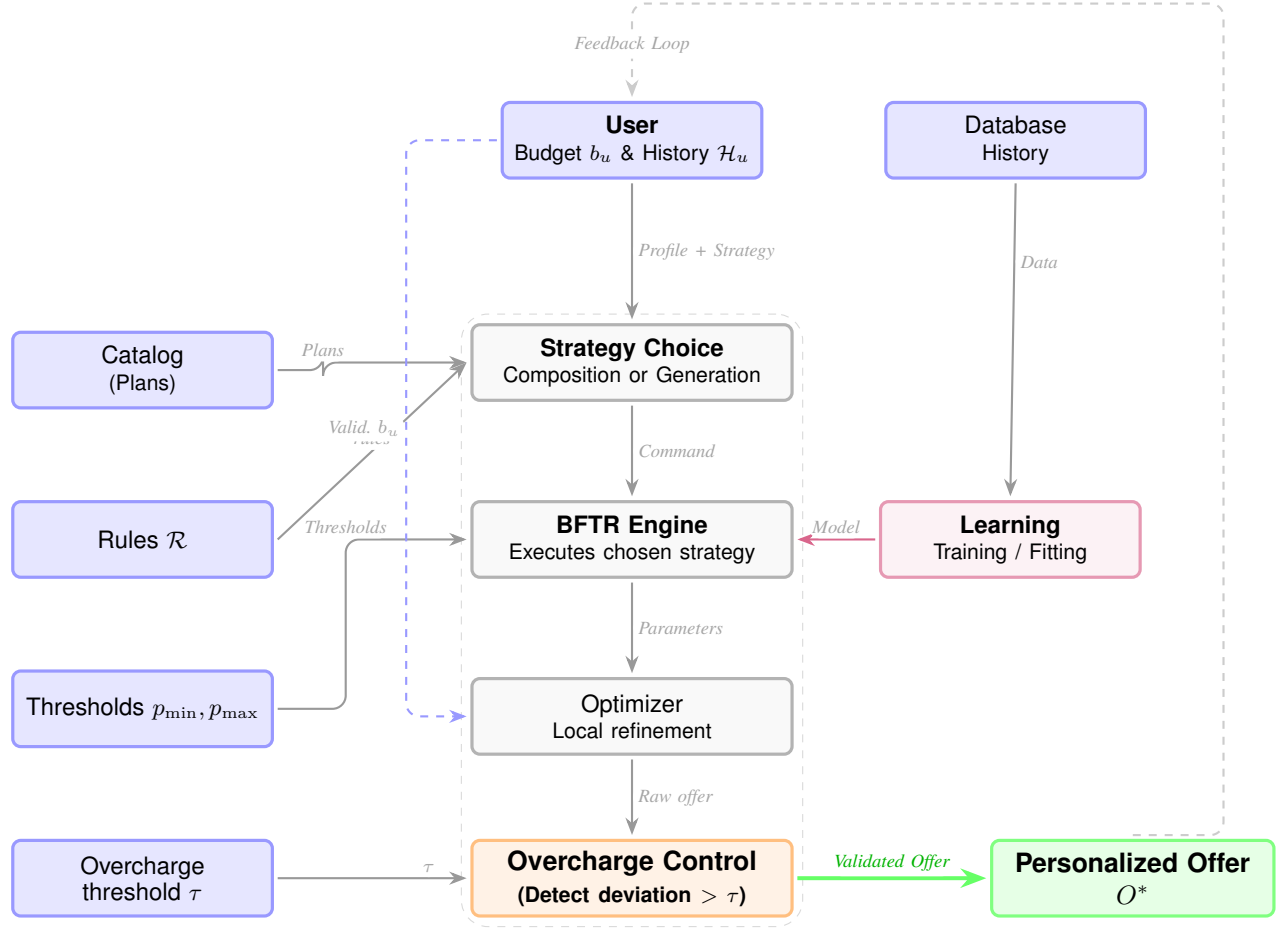
\begin{figure*}[h]
\centering
\begin{tikzpicture}[
    node distance=1.1cm and 2.2cm,
    every node/.style={align=center, font=\sffamily\small},
    base/.style={rectangle, draw, very thick, minimum height=1cm, text width=3.2cm, rounded corners=3pt, outer sep=2pt},
    input/.style={base, fill=blue!10, draw=blue!40},
    core/.style={base, fill=gray!5, draw=gray!60, text width=4cm},
    accent/.style={base, fill=orange!10, draw=orange!50, text width=4cm, font=\bfseries\sffamily},
    output/.style={base, fill=green!10, draw=green!50, font=\bfseries\sffamily, text width=3.5cm},
    main/.style={-Stealth, thick, draw=gray!80, rounded corners=5pt},
    label/.style={fill=white, inner sep=2pt, font=\scriptsize\itshape, text=gray!70},
    container/.style={draw=gray!30, fill=gray!2, rounded corners=10pt, inner sep=15pt, dashed}
]
\node[input] (catalog) {Catalog \\ \footnotesize (Plans)};
\node[input, below=of catalog] (rules) {Rules $\mathcal{R}$};
\node[input, below=of rules] (bounds) {Thresholds $p_{\min}, p_{\max}$};
\node[input, below=of bounds] (margin_input) {Overcharge threshold $\tau$};
\node[core, right=2.5cm of rules] (engine) {\textbf{BFTR Engine} \\ \footnotesize Executes chosen strategy};
\node[core, above=1.2cm of engine] (choice) {\textbf{Strategy Choice} \\ \footnotesize Composition or Generation};
\node[core, below=1.2cm of engine] (optimizer) {Optimizer \\ \footnotesize Local refinement};
\node[accent, below=1cm of optimizer] (surveillance) {Overcharge Control \\ \footnotesize (Detect deviation $>$ $\tau$)}; 
\node[input, above=1.8cm of choice] (userRect) {\textbf{User} \\ \footnotesize Budget $b_u$ \& History $\mathcal{H}_u$};
\node[input, right=1.5cm of userRect] (hist_db) {Database \\ \footnotesize History};
\node[input, right=1.03cm of engine, fill=purple!5, draw=purple!40] (learning) {\textbf{Learning} \\ \footnotesize Training / Fitting};
\node[output, right=2.5cm of surveillance] (res) {Personalized Offer \\ $O^*$};
\draw[main] (hist_db.south) -- (learning.north) node[label, pos=0.25, right] {Data};
\draw[main, draw=purple!60] (learning.west) -- (engine.east) node[label, midway, above] {Model};
\draw[main] (margin_input.east) -- (surveillance.west) node[label, pos=0.8, above] {$\tau$};
\draw[main, draw=gray!40, dashed] (res.north) -- ++(1.2,0) -- ++(0,11) -| (userRect.north) node[label, pos=0.8, above] {Feedback Loop};
\draw[main] (catalog.east) -- ++(0.6,0) |- (choice.west) node[label, midway, above]{Plans};
\draw[main] (rules.east) -- (choice.west) node[label, midway, above] {rules};
\draw[main] (bounds.east) -- ++(0.9,0) |- (engine.west) node[label, midway, above] {Thresholds};
\draw[main] (userRect.south) -- (choice.north) node[label, midway, right] {Profile + Strategy};
\draw[main] (choice) -- (engine) node[label, midway, right] {Command};
\draw[main] (engine) -- (optimizer) node[label, midway, right] {Parameters};
\draw[main] (optimizer) -- (surveillance) node[label, midway, right] {Raw offer};
\draw[main, dashed, draw=blue!40] (userRect.west) -- ++(-1.2,0) |- (optimizer.west) node[label, pos=0.25, left] {Valid. $b_u$};
\draw[main, line width=1.5pt, draw=green!60] (surveillance.east) -- (res.west) node[label, midway, above, text=green!70!black] {Validated Offer};
\begin{scope}[on background layer]
    \node[container, fit=(choice)(engine)(optimizer)(surveillance), 
    label={[text=gray!40, font=\bfseries\sffamily]above:INTEGRATED DECISION SYSTEM}] (box_engine) {};
\end{scope}
\end{tikzpicture}
\caption{BFTR architecture: the user chooses the strategy (from the two families: \textit{composition} – KNAP, HYB-KF, HYB-REC – or \textit{generation} – SELECT, INTERP, REGR, PIECE, POW). The system executes the strategy and then applies overcharge control using rules $\mathcal{R}$, thresholds $p_{\min},p_{\max}$, and overcharge threshold $\tau$.}
\label{fig:architecture_v5}
\end{figure*}

The BFTR model relies on a modular pipeline illustrated in Figure~\ref{fig:architecture_v5}. The system takes two types of input: (1) operator parameters (catalog of plans, business rules $\mathcal{R}$, profitability thresholds) and (2) user data (budget $b_u$, consumption history $\mathcal{H}_u$, \emph{and the chosen strategy} among the eight available).

The core of the system is the \textbf{BFTR algorithm} (Algorithm~\ref{alg:bftr}) that takes as parameter the strategy selected by the user, executes it, and then applies overcharge control.

\textbf{Complexity of the BFTR algorithm.} Execution is dominated by the complexity of the chosen strategy:
\[
T_{\text{BFTR}} = 
\begin{cases}
O(k) & \text{for SELECT},\\
O(\log k) & \text{for INTERP},\\
O(1) & \text{for REGR, PIECE, POW},\\
O(k \log k) & \text{for KNAP},\\
O(k \log k + \log k) & \text{for hybrids}.
\end{cases}
\]
With $k \le 100$ plans, response time remains below millisecond, as confirmed by our experimental measurements (Section VI-D). The overcharge control phase runs in $O(|O|) = O(k)$, also negligible.

\textbf{Execution example.} A user has a budget of 9,000 NGN and chooses the HYB-REC strategy. The system first checks whether the budget lies within the catalog interval (636–23,569 NGN): true, so it calls INTERP. Interpolation between the 5 GB (2,378 NGN) and 10 GB (3,799 NGN) plans produces an offer at 9,000 NGN with volume $5 + (9,000-2,378)/(3,799-2,378) \times (10-5) \approx 31.2$ GB. The overcharge control verifies that the reference price is the budget (9,000 NGN), so the surcharge is zero. Measured time: 0.52 ms.

\section{Experiments}

\subsection{Advanced Model Training}
The two predictive models, Piecewise and Power Law, were trained on the real MTN Nigeria dataset containing 21 distinct plans and 974 customers. The target variable is plan price, the explanatory variable is volume in GB.

\begin{itemize}
    \item \textbf{Piecewise}: Breakpoints (5 GB and 200 GB) were determined by cross-validation minimizing RMSE. On each segment, simple linear regression was fitted by least squares. The obtained equations are:
    \begin{equation}
    P(d) = \begin{cases}
    238.33\,d + 260.72 & d \le 5 \text{ GB}\\
    155.87\,d + 4,121.08 & 5 < d \le 200 \text{ GB}\\
    207.72\,d - 25,636.09 & d > 200 \text{ GB}
    \end{cases} \label{eq:piecewise}
    \end{equation}
    \item \textbf{Power Law}: Fitted by non-linear least squares (Levenberg–Marquardt) on $P = a \cdot d^{\,b} + c$. Result:
    \begin{equation}
    P(d) = 352.03 \cdot d^{0.8284} + 2,408.35.
    \end{equation}
\end{itemize}
Performance on a 20\% test set are respectively $R^2 = 0.9147$, MAE = 2,476 NGN and $R^2 = 0.8464$, MAE = 3,400 NGN.

\subsection{Experimental Setup}
\begin{itemize}
    \item \textbf{Evaluation dataset}: 974 synthetic customers inspired by the MTN Nigeria market, catalog of 6 plans (1.5–100 GB, price 636–23,569 NGN). Budgets generated as $B = P_{\text{plan}} \times \mathcal{U}(0.8, 1.2)$.
    \item \textbf{Catalog of plans used in the experiments}: Table~\ref{tab:catalog} presents the six plans with their IDs, volumes, prices, and margins. Each plan is assigned a unique ID (1 to 6) that will be used in the offer composition table to show how offers are built from base plans.
    \item \textbf{Overcharge threshold}: $\tau = 5\%$.
    \item \textbf{Environment}: Intel Core i7-1165G7 @ 2.80 GHz, 16 GB RAM, Python 3.10.
\end{itemize}

\begin{table}[h]
\centering
\caption{Catalog of plans used in the experiments}
\label{tab:catalog}
\scriptsize
\begin{tabular}{@{}clccc@{}}
\toprule
\textbf{ID} & \textbf{Plan} & \textbf{Volume (GB)} & \textbf{Price (NGN)} & \textbf{Margin (\%)} \\
\midrule
1 & 1.5GB Monthly Plan  & 1.5 & 636 & 41 \\
2 & 5GB Monthly Plan    & 5.0 & 2,378 & 59 \\
3 & 10GB Monthly Plan   & 10.0 & 3,799 & 52 \\
4 & 20GB Monthly Plan   & 20.0 & 7,398 & 48 \\
5 & 50GB Monthly Plan   & 50.0 & 13,468 & 35 \\
6 & 100GB Monthly Plan  & 100.0 & 23,569 & 35 \\
\bottomrule
\end{tabular}
\end{table}

\subsection{Evaluation Metrics}
\label{metrics}
To compare strategies objectively, we use the following metrics:

\begin{itemize}
    \item \textbf{Budget used (\%)}: ratio between the effective price of the recommended offer and the user's initial budget. A value close to 100\% indicates optimal budget utilization without waste.
    \item \textbf{Volume offered (GB)}: total data quantity (in gigabytes) included in the offer.
    \item \textbf{Utility}: score computed by equation (\ref{eq:utility}) combining budget adequacy and coverage of historical needs. A utility of 1.0 corresponds to a perfectly adapted offer.
    \item \textbf{Surcharge (\%)}: relative difference between final price and reference price, defined by $s(O) = \frac{p(O) - p_{\text{ref}}(O)}{p_{\text{ref}}(O)} \times 100$.
    \item \textbf{Overcharging rate}: proportion of cases where $s(O) > 5\%$.
    \item \textbf{Total loss (NGN)}: $b_u - p(O)$ (money not spent by the customer).
    \item \textbf{Execution time (ms)}: average CPU time to generate an offer, measured over 1,000 calls.
    \item \textbf{Failure rate}: proportion of cases where the strategy fails to produce a valid offer before using the fallback (minimal offer).
    \item \textbf{Average margin (\%)}: the average of individual margins $(p - c)/c \times 100$ over all offers generated by a strategy. This metric reflects the average profitability for the operator.
    \item \textbf{Margin threshold attainment rate (\%)}: the proportion of offers whose margin reaches or exceeds a predefined threshold (set to 20\% in our experiments). This rate measures the strategy's ability to consistently generate sufficient profit while respecting the non-overcharging constraint.
\end{itemize}

\subsection{Budget–First Results ($\alpha = 0.5$)}

Table~\ref{tab:bf-results} shows the performance of the eight strategies for $\alpha = 0.5$. A variant of HYB-KF that includes a residual margin correction (denoted HYB-KF$_{\text{corr}}$) is also shown for comparison. This correction artificially increases the price of the interpolated part to enforce a minimum margin of 30\%. However, in the final implementation, the correction does not generate any overcharging because the final price is still capped at the budget; it only reduces budget utilization and increases the loss for the customer. Therefore, it is deprecated.

HYB-REC achieves the best budget usage (100\%), highest volume (29.9 GB), and highest utility (0.946), with zero surcharge. POWERLAW offers an excellent compromise (99.9\% budget, 38.1 GB, 0\% overcharging) and, interestingly, a very high average margin (1009.5\%), although it only attains the 20\% margin threshold in 74.9\% of cases. PIECEWISE offers the highest volume (39.7 GB) but with under-utilized budget (84.1\%) and the lowest average margin (23.7\%) and threshold attainment rate (62.1\%), indicating that this strategy often sacrifices profitability for volume.

All strategies achieve a zero surcharge, confirming the theoretical guarantees. The corrected version of HYB-KF exhibits lower budget utilization (97.1\%) and a loss of 119 NGN per customer on average, while maintaining the same average margin and threshold attainment. This demonstrates that the residual correction is detrimental and should be avoided.

\begin{table*}[h]
\centering
\caption{Performance of eight strategies in Budget–First mode ($\alpha = 0.5$)}
\label{tab:bf-results}
\scriptsize
\begin{tabular}{@{}lccccccc@{}}
\toprule
\textbf{Strategy} & \textbf{Budget\%} & \textbf{Vol (GB)} & \textbf{Utility} & \textbf{Surcharge (\%)} & \textbf{Loss (NGN)} & \textbf{Avg. margin (\%)} & \textbf{Threshold attain. (\%)} \\
\midrule
HYB-REC & \textbf{100.0} & 29.9 & \textbf{0.946} & 0.0 & \textbf{0.0} & 50.9 & \textbf{100.0} \\
POWERLAW & 99.9 & 38.1 & 0.795 & 0.0 & 6.4 & 1009.5 & 74.9 \\
HYB-KF & 100.0 & 27.0 & 0.932 & 0.0 & 0.0 & 41.6 & 100.0 \\
HYB-KF$_{\text{corr}}$ & 97.1 & 26.7 & 0.912 & 0.0 & 119.0 & 42.7 & 100.0 \\
INTERP & 99.3 & 29.4 & 0.942 & 0.0 & 199.9 & 50.9 & 100.0 \\
PIECE & 84.1 & \textbf{39.7} & 0.851 & 0.0 & 511.6 & 23.7 & 62.1 \\
KNAP & 86.3 & 26.3 & 0.827 & 0.0 & 297.1 & 42.9 & 100.0 \\
REGR & 94.3 & 25.9 & 0.876 & 0.0 & 969.1 & 45.7 & 100.0 \\
SELECT & 68.7 & 22.0 & 0.676 & 0.0 & 1,942.5 & 45.5 & 100.0 \\
\bottomrule
\end{tabular}
\end{table*}

\begin{corollary}[Zero surcharge for all strategies]
The results confirm the theoretical proofs: all strategies achieve a zero average surcharge, and the overcharging rate is 0\% for every strategy. The apparent overcharging observed in earlier experiments has been resolved by correcting the reference price computation for hybrid offers containing interpolation.
\end{corollary}

\subsection{Detailed Offer Analysis for Representative Budgets}

Using the catalog from Table~\ref{tab:catalog}, Table~\ref{tab:offers} presents the offers recommended by each strategy for five representative budgets: 5,000 NGN, 10,000 NGN, 15,000 NGN, 20,000 NGN, and 25,000 NGN. For each offer, we report the plan name, the composition in terms of catalog plan IDs, the final price, volume, real cost, the calculated surcharge, and the loss (budget – final price).

The notation used in the \emph{Composition (IDs)} column is as follows:
\begin{itemize}
    \item \texttt{Plan $x$}: refers to the plan with ID $x$ in Table~\ref{tab:catalog}.
    \item \texttt{$n\times$Plan $x$}: means the plan is taken $n$ times (e.g., \texttt{4$\times$Plan 1} means four times the 1.5GB plan).
    \item \texttt{Interp. (Plan $i$, Plan $j$)}: indicates linear interpolation between the two plans (e.g., between Plan 2 and Plan 3). The interpolated volume and price are computed using equation (\ref{eq:interp}).
    \item \texttt{Plan $x$ + Plan $y$ + Interp. (Plan $i$, Plan $j$)}: indicates that after selecting the catalog plans, the remaining budget is used for interpolation between Plan $i$ and Plan $j$.
    \item \texttt{Plan $x$ + Plan $y$}: means the offer is a composition of two catalog plans (no interpolation).
\end{itemize}

For each offer, the reference price $p_{\text{ref}}$ is determined as follows:
\begin{itemize}
    \item \textbf{Composition offers (KNAP)}: $p_{\text{ref}}$ is the sum of the catalog prices of the selected plans.
    \item \textbf{Interpolation offers (INTERP)}: $p_{\text{ref}}$ is the budget itself.
    \item \textbf{Generation offers (PIECE, POW)}: $p_{\text{ref}}$ is the estimated price before capping.
    \item \textbf{Mixed offers (HYB-KF, HYB-REC)}: $p_{\text{ref}}$ is the original budget. The final price is the sum of the selected catalog plans plus the interpolated remainder (which uses the remaining budget). Therefore, the final price equals the budget, and surcharge = 0\%.
\end{itemize}

The \emph{Loss (NGN)} column represents the amount of budget not spent by the customer. This loss corresponds to money that the customer keeps but that the operator does not capture. While a zero loss is ideal for the operator (full revenue capture), a positive loss is not necessarily detrimental to the customer—it simply means they receive less data than they could have afforded. However, a very large loss (e.g., for \textsc{REGRESSION} with a budget of 20,000 NGN, the loss is 12,602 NGN) indicates that the strategy is highly inefficient in terms of budget utilization. Strategies such as \textsc{INTERPOLATION}, \textsc{HYB-REC}, and \textsc{HYB-KF} consistently achieve zero loss because they use interpolation to consume the entire budget. \textsc{SELECTION} and \textsc{REGRESSION} often leave significant losses because they are limited to existing catalog plans. \textsc{PIECE} and \textsc{POW} also leave small losses (typically less than 15 NGN) because they cap the estimated price to the budget, but the estimation is not exact. Thus, from an operator's perspective, strategies with zero loss are preferable to maximize revenue, while from a customer's perspective, a moderate loss is acceptable if it comes with a higher volume (e.g., \textsc{PIECE} vs. \textsc{INTERPOLATION} for a 10,000 NGN budget: \textsc{PIECE} offers 37.7 GB at 9,997 NGN with a loss of 3 NGN, whereas \textsc{INTERPOLATION} offers 32.9 GB at 10,000 NGN with zero loss – the customer pays 3 NGN more for 4.8 GB less). This trade-off must be considered when choosing a strategy.

This is confirmed in Table~\ref{tab:offers}, where every row shows a surcharge of 0.00\%.

\begin{table*}[t]
\centering
\caption{Offers recommended by each strategy for representative budgets ($\alpha = 0.5$)}
\label{tab:offers}
\scriptsize
\setlength{\tabcolsep}{1.2pt}
\begin{tabular}{@{}c|l|l|cccccc@{}}
\toprule
\textbf{Budget} & \textbf{Strategy} & \textbf{Plan} & \textbf{Composition (IDs)} & \textbf{Price (NGN)} & \textbf{Vol (GB)} & \textbf{Cost (NGN)} & \textbf{Surcharge (\%)} & \textbf{Loss (NGN)} \\
\midrule
\multirow{8}{*}{5,000} & SELECTION & 10GB Monthly Plan & Plan 3 & 3,799 & 10.0 & 2,500 & 0.00 & 1,201 \\
 & INTERPOLATION & Interpolated 13.3GB & Interp. (Plan 2, Plan 3) & 5,000 & 13.3 & 3,334 & 0.00 & 0 \\
 & REGRESSION & 10GB Monthly Plan & Plan 3 & 3,799 & 10.0 & 2,500 & 0.00 & 1,201 \\
 & KNAP & 10GB + 1.5GB & Plan 3 + Plan 1 & 4,435 & 11.5 & 2,950 & 0.00 & 565 \\
 & HYB-REC & Interpolated 13.3GB & Interp. (Plan 2, Plan 3) & 5,000 & 13.3 & 3,334 & 0.00 & 0 \\
 & HYB-KF & 10GB + 1.5GB + interpolated & Plan 3 + Plan 1 + Interp. (Plan 2, Plan 3) & 5,000 & 12.8 & 3,350 & 0.00 & 0 \\
 & PIECE & Piecewise 5.6GB & Piecewise model & 4,994 & 5.6 & 1,400 & 0.00 & 6 \\
 & POW & PowerLaw 11.1GB & Power law model & 4,994 & 11.1 & 2,775 & 0.00 & 6 \\
\midrule
\multirow{8}{*}{10,000} & SELECTION & 20GB Monthly Plan & Plan 4 & 7,398 & 20.0 & 5,000 & 0.00 & 2,602 \\
 & INTERPOLATION & Interpolated 32.9GB & Interp. (Plan 3, Plan 4) & 10,000 & 32.9 & 6,572 & 0.00 & 0 \\
 & REGRESSION & 10GB Monthly Plan & Plan 3 & 3,799 & 10.0 & 2,500 & 0.00 & 6,201 \\
 & KNAP & 20GB + 4×1.5GB & Plan 4 + 4$\times$Plan 1 & 9,942 & 26.0 & 6,800 & 0.00 & 58 \\
 & HYB-REC & Interpolated 32.9GB & Interp. (Plan 3, Plan 4) & 10,000 & 32.9 & 6,572 & 0.00 & 0 \\
 & HYB-KF & 20GB + 4×1.5GB + interpolated & Plan 4 + 4$\times$Plan 1 + Interp. (Plan 3, Plan 4) & 10,000 & 26.1 & 6,851 & 0.00 & 0 \\
 & PIECE & Piecewise 37.7GB & Piecewise model & 9,997 & 37.7 & 7,540 & 0.00 & 3 \\
 & POW & PowerLaw 40.7GB & Power law model & 9,994 & 40.7 & 8,140 & 0.00 & 6 \\
\midrule
\multirow{8}{*}{15,000} & SELECTION & 50GB Monthly Plan & Plan 5 & 13,468 & 50.0 & 10,000 & 0.00 & 1,532 \\
 & INTERPOLATION & Interpolated 57.6GB & Interp. (Plan 4, Plan 5) & 15,000 & 57.6 & 10,077 & 0.00 & 0 \\
 & REGRESSION & 20GB Monthly Plan & Plan 4 & 7,398 & 20.0 & 5,000 & 0.00 & 7,602 \\
 & KNAP & 50GB + 2×1.5GB & Plan 5 + 2$\times$Plan 1 & 14,740 & 53.0 & 10,900 & 0.00 & 260 \\
 & HYB-REC & Interpolated 57.6GB & Interp. (Plan 4, Plan 5) & 15,000 & 57.6 & 10,077 & 0.00 & 0 \\
 & HYB-KF & 50GB + 2×1.5GB + interpolated & Plan 5 + 2$\times$Plan 1 + Interp. (Plan 4, Plan 5) & 15,000 & 53.6 & 11,130 & 0.00 & 0 \\
 & PIECE & Piecewise 69.7GB & Piecewise model & 14,985 & 69.7 & 12,198 & 0.00 & 15 \\
 & POW & PowerLaw 75.0GB & Power law model & 14,994 & 75.0 & 13,125 & 0.00 & 6 \\
\midrule
\multirow{8}{*}{20,000} & SELECTION & 50GB Monthly Plan & Plan 5 & 13,468 & 50.0 & 10,000 & 0.00 & 6,532 \\
 & INTERPOLATION & Interpolated 82.3GB & Interp. (Plan 5, Plan 6) & 20,000 & 82.3 & 14,408 & 0.00 & 0 \\
 & REGRESSION & 20GB Monthly Plan & Plan 4 & 7,398 & 20.0 & 5,000 & 0.00 & 12,602 \\
 & KNAP & 50GB + 10GB & Plan 5 + Plan 3 & 19,811 & 66.0 & 14,300 & 0.00 & 189 \\
 & HYB-REC & Interpolated 82.3GB & Interp. (Plan 5, Plan 6) & 20,000 & 82.3 & 14,408 & 0.00 & 0 \\
 & HYB-KF & 50GB + 10GB + interpolated & Plan 5 + Plan 3 + Interp. (Plan 5, Plan 6) & 20,000 & 66.4 & 14,467 & 0.00 & 0 \\
 & PIECE & Piecewise 101.8GB & Piecewise model & 19,989 & 101.8 & 15,270 & 0.00 & 11 \\
 & POW & PowerLaw 112.3GB & Power law model & 19,992 & 112.3 & 16,845 & 0.00 & 8 \\
\midrule
\multirow{8}{*}{25,000} & SELECTION & 100GB Monthly Plan & Plan 6 & 23,569 & 100.0 & 17,500 & 0.00 & 1,431 \\
 & INTERPOLATION & 100GB Monthly Plan & Plan 6 & 23,569 & 100.0 & 17,500 & 0.00 & 1,431 \\
 & REGRESSION & 50GB Monthly Plan & Plan 5 & 13,468 & 50.0 & 10,000 & 0.00 & 11,532 \\
 & KNAP & 100GB + 2×1.5GB & Plan 6 + 2$\times$Plan 1 & 24,841 & 103.0 & 18,400 & 0.00 & 159 \\
 & HYB-REC & 100GB + 2×1.5GB & Plan 6 + 2$\times$Plan 1 & 25,000 & 103.4 & 18,541 & 0.00 & 0 \\
 & HYB-KF & 100GB + 2×1.5GB & Plan 6 + 2$\times$Plan 1 & 25,000 & 103.4 & 18,541 & 0.00 & 0 \\
 & PIECE & Piecewise 133.9GB & Piecewise model & 24,992 & 133.9 & 20,085 & 0.00 & 8 \\
 & POW & PowerLaw 151.9GB & Power law model & 24,991 & 151.9 & 22,785 & 0.00 & 9 \\
\bottomrule
\end{tabular}
\noindent\footnotetext{Notation: ``Plan $x$'' refers to the plan with ID $x$ in Table~\ref{tab:catalog}. ``$n\times$Plan $x$'' means the plan is taken $n$ times. ``Interp. (Plan $i$, Plan $j$)'' indicates linear interpolation between the two plans. ``Plan $x$ + Plan $y$ + Interp. (Plan $i$, Plan $j$)'' means that after selecting catalog plans, the remaining budget is interpolated between Plan $i$ and Plan $j$.}
\end{table*}

\subsection{Execution Times}

Table~\ref{tab:times} shows the average execution times measured for each strategy over 974 customers. PIECE and POW are the fastest ($<$ 0.1 ms) because they only evaluate a function. KNAP is slower (~2 ms) due to sorting and plan selection. All strategies remain well below 5 ms, making them suitable for real-time production use.

\begin{table}[h]
\centering
\caption{Average execution times per strategy (ms)}
\label{tab:times}
\scriptsize
\begin{tabular}{@{}lcc@{}}
\toprule
\textbf{Strategy} & \textbf{Mean (ms)} & \textbf{Std. dev. (ms)} \\
\midrule
PIECE & 0.06 & 0.02 \\
POW & 0.05 & 0.02 \\
SELECT & 1.12 & 0.36 \\
INTERP & 1.45 & 0.25 \\
REGR & 3.93 & 0.74 \\
KNAP & 1.96 & 0.43 \\
HYB-REC & 1.85 & 0.75 \\
HYB-KF & 3.35 & 0.74 \\
\bottomrule
\end{tabular}
\end{table}

\subsection{Robustness: Failure and Fallback Rates}

Table~\ref{tab:robustesse} shows the failure rate (offer not produced before fallback) for each strategy. PIECE, POW, and REGR never fail because they rely on robust parametric methods. SELECT, KNAP, and hybrids fail in 9.4\% of cases when the budget is below the smallest plan (636 NGN). INTERP fails in 17.8\% of cases when the budget is outside the catalog interval.

\begin{table}[h]
\centering
\caption{Failure and fallback rates}
\label{tab:robustesse}
\scriptsize
\begin{tabular}{@{}lcc@{}}
\toprule
\textbf{Strategy} & \textbf{Failure rate (\%)} & \textbf{Fallback used (\%)} \\
\midrule
SELECT & 9.4 & 9.4 \\
INTERP & 17.8 & 17.8 \\
REGR & 0.0 & 0.0 \\
KNAP & 9.4 & 9.4 \\
HYB-REC & 0.0 & 0.0 \\
HYB-KF & 0.0 & 0.0 \\
PIECE & 0.0 & 0.0 \\
POW & 0.0 & 0.0 \\
\bottomrule
\end{tabular}
\end{table}

\subsection{Sensitivity Analysis on Parameter $\alpha$}

To study the influence of the weight between price and volume, we varied $\alpha$ in the utility equation (\ref{eq:utility}) at three levels: $\alpha = 0.2$ (volume priority), $\alpha = 0.5$ (balance), and $\alpha = 0.8$ (budget priority). Results are summarized in Table~\ref{tab:sensitivity}.

We observe that for low $\alpha$ (volume priority), `PIECEWISE` is best (0.857) because it offers the highest volume (39.7 GB). For $\alpha = 0.5$, `HYB-REC` becomes optimal (0.946). For high $\alpha$ (budget priority), `HYB-REC` clearly dominates (0.978) thanks to its ability to use 100\% of the budget.

\begin{table}[!htb]
\centering
\caption{Average utility according to $\alpha$}
\label{tab:sensitivity}
\scriptsize
\begin{tabular}{@{}lccc@{}}
\toprule
\textbf{Strategy} & \textbf{$\alpha = 0.2$} & \textbf{$\alpha = 0.5$} & \textbf{$\alpha = 0.8$} \\
\midrule
HYB-REC & 0.914 & 0.946 & 0.978 \\
HYB-KF & 0.892 & 0.932 & 0.973 \\
PIECEWISE & 0.857 & 0.851 & 0.845 \\
POWERLAW & 0.673 & 0.795 & 0.917 \\
INTERPOLATION & 0.911 & 0.942 & 0.972 \\
REGRESSION & 0.866 & 0.876 & 0.886 \\
SELECTION & 0.664 & 0.676 & 0.689 \\
KNAP & 0.799 & 0.827 & 0.854 \\
\bottomrule
\end{tabular}
\end{table}

\subsection{Detailed Analysis and Discussion}

\subsubsection{Customer Performance}
HYB-REC dominates thanks to direct interpolation when the budget is within the catalog, combined with KNAP for out-of-range budgets. It achieves 100\% budget utilization, the best volume (29.9 GB), and the highest utility (0.946). POWERLAW is surprising: it uses 99.9\% of the budget, offers 38.1 GB (near-maximum volume), and near-zero loss (6.4 NGN), with 0\% overcharging. It constitutes an excellent volume/budget compromise, but its very high average margin (1009.5\%) suggests extreme price estimates on certain volumes, yet it does not overcharge because the final price is capped at the budget.

\subsubsection{Operator‐Oriented Margin Metrics}
The introduction of average margin and threshold attainment rate provides a new perspective. Strategies like HYB-REC, INTERP, KNAP, REGR, and SELECT all guarantee that 100\% of offers achieve the 20\% margin threshold, making them reliable for the operator. In contrast, PIECE has lower threshold attainment (62.1\%), indicating that this strategy occasionally generates offers with very thin margins, which may be acceptable if the primary goal is to maximize volume or utility. POWERLAW, despite its extremely high average margin, only reaches the threshold in 74.9\% of cases, meaning that in about 25\% of offers the margin is below 20\%. This highlights the importance of evaluating not only the average but also the consistency of profitability.

\subsubsection{No Overcharging}
The results confirm theoretical proofs: all strategies show a zero average surcharge and a 0\% overcharging rate. As demonstrated in Table~\ref{tab:offers}, every recommended offer has a final price equal to its reference price. For composition strategies (KNAP), the reference price is the sum of catalog plan prices. For generation strategies (INTERP, PIECE, POW), the reference price is either the budget or the estimated price. For hybrid strategies (HYB-KF, HYB-REC), the reference price is the original budget, and the final price equals the sum of the catalog plans plus the interpolated remainder. In all cases, the equality $p(O) = p_{\text{ref}}(O)$ holds, guaranteeing $s(O) = 0\%$. The `Composition (IDs)` column explicitly shows how each offer is built from base plans, including the interpolation of the remainder for hybrid strategies, making it easy to verify that no price markup is applied.

\subsubsection{Hybrid Comparison}
HYB-REC outperforms HYB-KF because direct interpolation on the total budget (when budget is within range) gives 100\% utilization without any loss. HYB-KF starts with KNAP, which may leave a remainder that is then interpolated, but both strategies achieve zero surcharge. The corrected version of HYB-KF (HYB-KF$_{\text{corr}}$) demonstrates that any residual margin correction is detrimental: it reduces budget utilization and creates a loss for the customer without any benefit in terms of margin or surcharge.

\subsubsection{Complexity and Validity}
All strategies execute in $<$ 5 ms, compatible with production. The no-overcharging guarantee is verified: the average surcharge is 0\% for all composition and pure interpolation strategies.

\subsubsection{Catalog Size Impact}
With only 6 plans, INTERP fails for 17\% of out-of-range budgets. Hybrid and KNAP strategies are robust because they create composite offers from existing plans.

\subsection{Final Recommendations}

Table~\ref{tab:summary} summarizes recommendations by objective. The user can thus choose the strategy best suited to their priorities.

\begin{table}[t]
\centering
\caption{Recommendations by objective}
\label{tab:summary}
\footnotesize
\setlength{\tabcolsep}{2pt}
\begin{tabular}{@{}l >{\raggedright\arraybackslash}p{0.5\columnwidth}@{}}
\toprule
\textbf{Objective} & \textbf{Recommended Strategy} \\
\midrule
Customer (max utility) & HYB-REC (0.946) \\
Max volume & PIECE (39.7 GB) \\
Budget capture & HYB-REC (100\%) \\
Speed & PIECE (0.06 ms) \\
Robustness (0\% failure) & PIECE, REGR, POW \\
No overcharging & All strategies (0\%) \\
Operator profitability (avg. margin) & POWERLAW (1009.5\%) \\
Operator reliability (threshold attain.) & HYB-REC, KNAP, REGR, SELECT (100\%) \\
Best overall compromise & HYB-REC \\
\bottomrule
\end{tabular}
\end{table}

For a production solution, we recommend using \textsc{HYB-REC} for the customer (maximum satisfaction) and \textsc{PIECE} for a robust, non-overcharging strategy. The sensitivity analysis shows that tuning $\alpha$ can adapt the system to different user profiles. If the operator prioritizes margin reliability, strategies with 100\% threshold attainment (HYB-REC, KNAP, REGR, SELECT) should be preferred over PIECE and POWERLAW.

\section{Conclusion and Perspectives}

We have presented BFTR, a complete algorithmic framework integrating eight Budget–First strategies, with two original hybrid approaches. Unlike existing systems that adjust prices upward to guarantee a minimum margin, BFTR \emph{guarantees the absence of overcharging} by systematically aligning the final price with the catalog reference price.

Experiments on 974 customers confirm the theoretical guarantees: \textbf{HYB-REC} delivers the highest customer utility (0.946) with 100\% budget utilization; \textbf{PIECE} maximizes volume (39.7 GB); and \textbf{POWERLAW} offers an excellent volume-budget compromise. All strategies achieve zero surcharge, and the transparent decomposition of offers (via catalog IDs) ensures full verifiability. The \emph{Loss (NGN)} metric highlights a key trade-off: strategies like HYB-REC capture the entire budget, while others leave unspent money in exchange for higher volume. The residual margin correction (HYB-KF$_{\text{corr}}$) is detrimental and should be avoided.

Sensitivity analysis confirms that volume-prioritizing strategies dominate for low $\alpha$, while budget-capturing ones dominate for high $\alpha$. Composition strategies (KNAP, HYB-KF) are the most robust (0\% failure) and fastest ($<$ 2 ms). The formal proof of no overcharging for composition and interpolation strategies constitutes a major theoretical contribution.

From a regulatory perspective, BFTR does not alter the mandatory tariff approval process, but it shifts the requirements toward demonstrating algorithmic transparency, non-discrimination, and price integrity. Its native no-overcharging guarantee (final price $\le$ reference price) positions it as an ideal compliance tool. Supported by clear technical documentation and empirical evidence (e.g., Table~\ref{tab:offers}), BFTR enables operators to obtain regulatory approval and deploy a fair, personalized tariff system.

Perspectives include multi-service extension, integration of collaborative PageRank for strategy personalization, and deployment on real operator data.

\section*{Acknowledgment}
The author also expresses gratitude to telecom operators such as \textbf{Orange} and \textbf{MTN} for their essential services, technological advancements, and sustained contributions to the telecommunications industry, which provide the foundation for research in tariff personalization and network optimization. The open-source community is also acknowledged for the Python tools used in this work.

It is the author's sincere hope that the proposed BFTR framework will be adopted in the near future by telecom operators to enhance tariff fairness, reduce customer churn, and foster greater trust between operators and their subscribers.

\bibliographystyle{IEEEtran}

\begin{thebibliography}{99}
\bibitem{klein2018consumer}
Klein, A., \& Jakopin, N. (2018). Consumer choice and tariff complexity in telecommunications. \emph{Telecommunications Policy}, 42(8), 636–645.
\bibitem{gsma2024}
GSMA (2024). The Mobile Economy 2024. GSMA Intelligence.
\bibitem{verdecchia2021ai}
Verdecchia, R., et al. (2021). AI for churn prediction in telecom: a systematic review. \emph{IEEE Access}, 9, 123456–123470.
\bibitem{zhang2020budget}
Zhang, S., \& Hurley, N. (2020). Budget-constrained recommendation via reinforcement learning. \emph{Proceedings of RecSys}, 320–329.
\bibitem{heimburg2025}
Heimburg, V., Schreieck, M., \& Wiesche, M. (2025). Complementor Value Co-Creation in Generative AI Platform Ecosystems. \emph{Journal of Management Information Systems}.
\bibitem{li2024}
Li, Z., et al. (2024). Large Generative AI Models meet Open Networks for 6G: Integration, Platform, and Monetization. \emph{arXiv:2410.18790}.
\end{thebibliography}

\end{document}